\documentclass[10pt,twocolumn,letterpaper]{article}

\usepackage{iccv}
\usepackage{times}
\usepackage{epsfig}
\usepackage{graphicx}
\usepackage{amsmath}
\usepackage{amssymb}
\usepackage{cite}

\usepackage[breaklinks=true,bookmarks=false]{hyperref}

\iccvfinalcopy 

\def\iccvPaperID{****} 

\ificcvfinal\fi

\begin{document}

\title{Comprehensive Video Understanding: Video summarization with content-based video recommender design}

\author{
Yudong Jiang\textsuperscript{1}, Kaixu Cui\textsuperscript{1}, Bo Peng\textsuperscript{2}\footnotemark[3], Changliang Xu\textsuperscript{1} \\
\textsuperscript{1}Xinhua Zhiyun Technology Co., Ltd. \\
\textsuperscript{2}Dept. of Statistics, Columbia University in the City of New York  \\
{\tt\small \{jiangyudong,cuikaixu,xuchangliang\}@shuwen.com, bp2548@columbia.edu}
}

\maketitle
\ificcvfinal\thispagestyle{empty}\fi

\renewcommand{\thefootnote}{\fnsymbol{footnote}} 
\footnotetext[3]{Bo Peng is an intern in Xinhua Zhiyun Technology Co., Ltd. when doing this work.}

\begin{abstract}
   Video summarization aims to extract keyframes/shots from a long video. Previous methods mainly take diversity and representativeness of generated summaries as prior knowledge in algorithm design. In this paper, we formulate video summarization as a content-based recommender problem, which should distill the most useful content from a long video for users who suffer from information overload.  A scalable deep neural network is proposed on predicting if one video segment is a useful segment for users by explicitly modelling both segment and video. Moreover, we accomplish scene and action recognition in untrimmed videos in order to find more correlations among different aspects of video understanding tasks. Also, our paper will discuss the effect of audio and visual features in summarization task. We also extend our work by data augmentation and multi-task learning for preventing the model from early-stage overfitting. The final results of our model win the first place in ICCV 2019 CoView Workshop Challenge Track.
\end{abstract}

\section{Introduction}
\begin{figure}[t]
\begin{center}
   \includegraphics[width=1.0\linewidth]{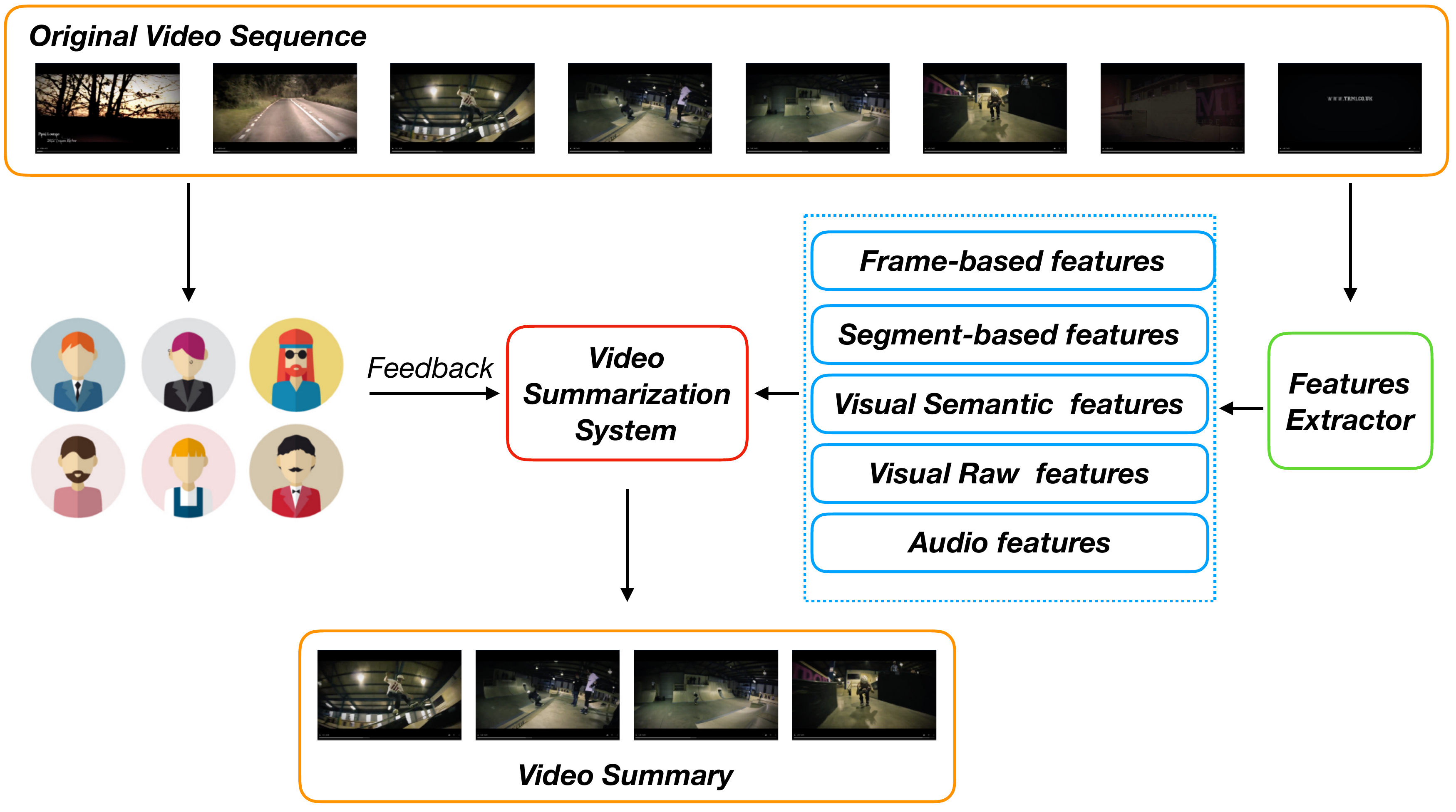}
\end{center}
   \caption{The whole picture of our video summarization system. Feature extractor can be a model trained on different video understanding tasks. Summarization system use viewers' feedback and features from different source to generate the video summary. Users' feedback is treated as supervised signals but not a necessary condition in the prediction process.}
\label{fig:long}
\label{fig:onecol}
\end{figure}
As information overload becomes a more and more serious topic in modern society, many efficient tools have been designed to overcome “information anxiety”.  Videos, the fastest-growing information carrier, will account for more than 80\% of all Internet traffic by 2020~\cite{SVA2018}. Video summarization aims to address this problem by extracting keyframes/shots from a long video, which contains the most useful content for people. It serves as a vital way of comprehensively understanding video data in research area while saving viewers' time on information acquisition. This technology also gains the attention of both the academic world and the industry. Adobe has already presented video summarization feature in their products for video editing. Some cloud computing companies, like Microsoft Azure, AliCloud, provide this function as an online service.   
   
One of the main challenges in video summarization is subjectivity, different people may have different selections on key shots for a same video. We investigate two top referenced dataset SumMe~\cite{gygli2014creating} and  TVSum~\cite{song2015tvsum}, which both provide manually curated consistency benchmark, with F1 score of 0.31 for SumMe, F1 score of 0.36 for TVSum. Based on the assumptions above, we don't take diversity or representativeness and other subjective attributes into consideration, instead, we focuse on solely the key shots and the full video. The mission of our algorithm is to find the most attractive video segments for a group of annotators/users when they watch a long video.

Content-based recommender modelling is one of the most important technology in recommendation system, especially for alleviating cold start problem, which recommend items with similar content to what  users like. It is usually formalized as a similarity learning problem~\cite{lee2017large}. A content-based video recommender deep model learns compact representation of videos and constructs a bridge between users’ feedback and video semantics information. We develop and deploy the HighlightNet to learn annotators’ preferences, the model is supervised by the segment’s importance feedbacks. It can easily combine different inputs, such as raw features, high-level features, audio features and vision feature together under this framework. The holistic picture of our work is shown in figure 1.

Currently, as video summarization problem appeals to a lot of researchers, many state-of-the-art approaches have been presented in solving this problem~\cite{SVA2018,zhang2016video,zhang2018retrospective,zhang2019dtr}. Generally, they treat video summarization as sequence-to-sequence learning problem. Since RNN and its variants LSTM, GRU are very efficient in modelling long time dependencies under encoder-decoder architectures, many works in machine translation, image/video captioning, reading comprehension adopt these technologies. In this paper, GRU modelling is adopted in order to consider both the isolated segment, and its roles in the whole video.

Video summarization, as one of the comprehensive video understanding tasks, is extremely difficult and requires a large amount of data in deep learning architecture. However, the collection of such summarization labels is time-consuming and labor-intensive, resulting in an insufficient dataset. Since supervised learning can easily overfit on small data, many methods have been explored on alleviating this issue. Many state-of-the-art videos summarization work address this problem by unsupervised learning, semi-supervised learning, or multi-task learning~\cite{zhang2018retrospective,zhang2019dtr,zhou2018reinforcement}. We explore the self-supervised learning on video sequence modelling and joint training with segments importance score prediction. Another way to confront overfitting is to shuffle information flows which potentially improves the model's generalization ability.
The contribution of our work:

   [1] {\bf We unify various inputs on different semantic levels to one framework by formatting video summarisation into a recommender problem.}

   [2] {\bf We develop an algorithm that models independent segments and segment sequence(whole video).}

   [3] {\bf We extend the summarisation framework with self-supervised learning and data augmentation to deal with lack of labelled data.}

\section{Related Work}
Video classification, as a fundamental task for video understanding, has been studied seriously. Many high-quality datasets~\cite{abu2016youtube,kay2017kinetics,kuehne2011hmdb,soomro2012ucf101}are published which drives the research to a higher level. Supervised learning brings the performance of algorithms nearly to human performance on large-scale video classification tasks~\cite{feichtenhofer2018slowfast,he2016resnet,karpathy2014large,szegedy2015going,tran2018closer,wang2016temporal,wang2018non}.

A significant number of deep learning based frameworks have been explored recently in solving video summarization~\cite{SVA2018,wei2018SAN,zhang2016video,zhang2019dtr,zhou2018reinforcement}. K. Zhang et al. creatively applied LSTM in supervised video sequence labelling to model video temporal information with good performance~\cite{zhang2016video}. Jiri Fajtl et.al introduced self-attention instead of RNN to enhance computation efficiency~\cite{SVA2018}. K. Zhou et al. showed that fully unsupervised learning can outperform many supervised methods by considering diversity and representativeness in reinforcement learning-based framework~\cite{zhou2018reinforcement}. Y. Zhang et al. introduced adversarial loss to video summarization which learns a dilated temporal relational generator and a discriminator with three-player loss~\cite{zhang2019dtr}.

Researchers believe learning good visual representation can help deep neural network improve both fitting and generalization ability, especially in classification task~\cite{fernando2017self,lee2017unsupervised,misra2016shuffle,srivastava2015unsupervised}. Some video summarization work adopts unsupervised learning as an auxiliary task to help to improve the supervised learning system’s performance. K. Zhang et al. used “retrospective encoder” that embeds the predicted summary and original video into the same abstract semantic space with closer distance. The semi-supervised settings help increase F-score on TVSum dataset from 63.9\% to 65.2\%~\cite{zhang2018retrospective}. K. Zhou et al. on the other hand, extending their reinforcement learning based framework to semi-supervised style elevated performance on both SumMe and TVSum comparing to purely supervised or unsupervised methods~\cite{zhou2018reinforcement}. 

For the subjectivity of video summarization task, Kanehira et al. provided a solution by building a summary that depends on the particular aspect of a video the viewer focuses on~\cite{kanehira2018aware}. Joonseok Lee et.al from Google published a content-only video recommendation system which we regarded as a work close to video summarization~\cite{lee2017large}.

\section{Approach}
In this section, we detail our method on CoView 2019 comprehensive video understanding challenge track. First, we simply introduce our work on action and scene recognition in untrimmed video. Second, we formalize the summarization problem and present our solution based on deep neural network. Third, we talk about some important technology on how to prevent the DNN from early-stage overfitting.

\subsection{Action and Scene Recognition in Untrimmed Video}
We first adopt I3D with non-local blocks video classification~\cite{carreira2017quo,wang2018non} in this subtask. A ResNet101 backbone is used for this framework. We tried two ways of training: 

\textit{1. Action and scene recognition task sharing the same backbone with two SoftMax loss branches for each classification tasks. }

\textit{2. We get one model each for action and scene recognition.}

Since the training data just comes from only 1000 videos, the scene data for training is too small and lacks variation for recognition. There are many public image scene classification datasets such as Place360~\cite{zhou2017places} and SUN~\cite{xiao2010sun} for image scene recognition. We assume the video scene classification task can be trained by image classification without much information loss. We can easily make full use of public datasets to potentially improve models’ robustness. We use average pooling operation on image classification predictions for final video level results. To capture the advantages of both video-based and image-based models, we ensemble the output scores of two models together with logistic regression. 

\subsection{Video Summarization Framework} 
We take the video summarization task as a content-based recommender problem. Let $$  V_{ m }=\left[s_{ 1 },s_{ 2 },s_{ 3 },...,s_{ T_{ N } } \right]^{ T }\in {  \mathbb{R} }^{ T_ {N} \times sd },  $$ be the feature matrix for a video m with $T_{N}$ segments in $sd$ dimension.

\begin{figure}[t]
\begin{center}
   \includegraphics[width=1.0\linewidth]{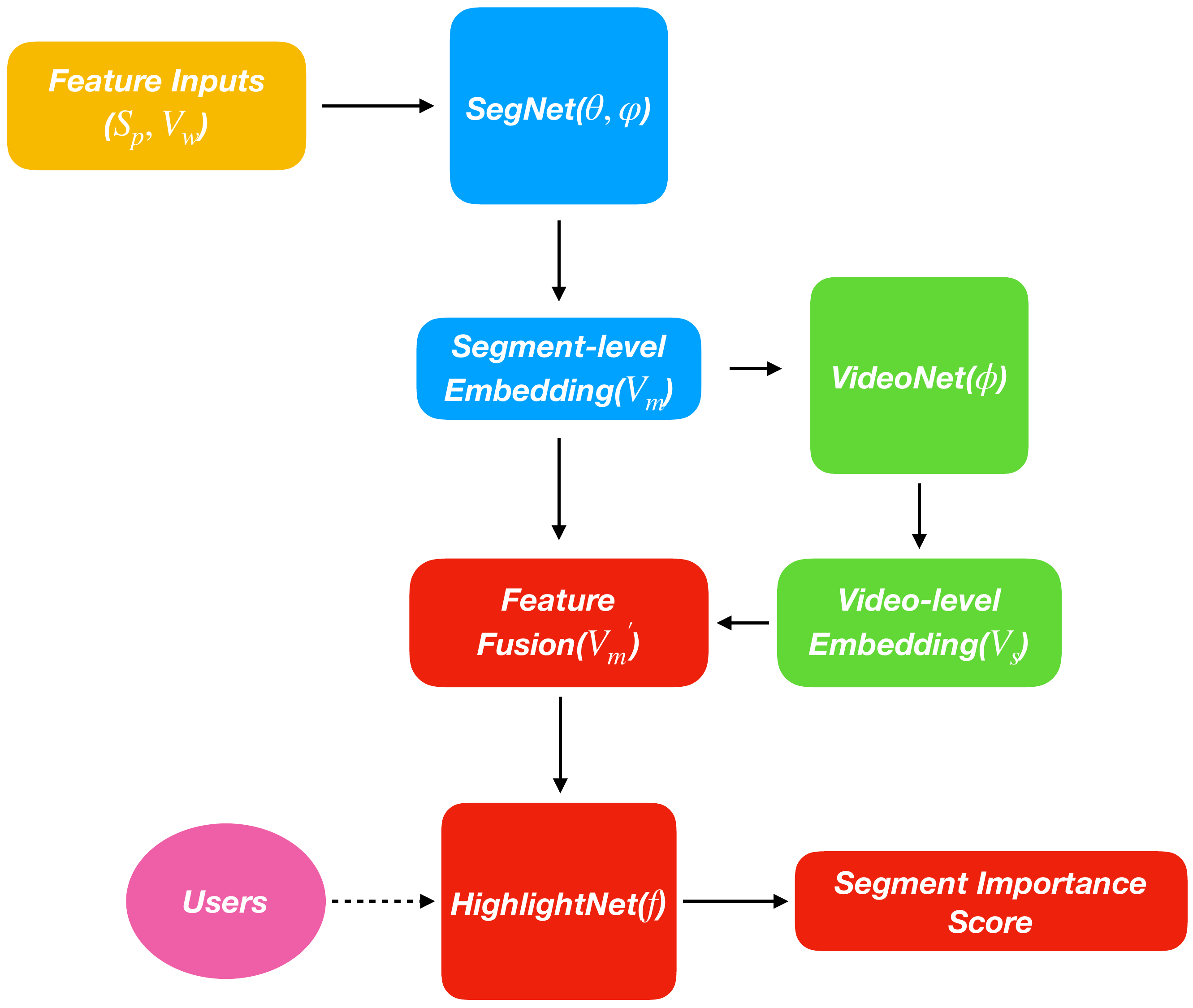}
\end{center}
   \caption{The Video Summarization Network structure. It contains three subnetworks including: SegNet, VideoNet, and HighlightNet.}
\label{fig:long}
\label{fig:onecol}
\end{figure}

For each segment we have a mean importance score $a_{i}$ as users' feedback, can be written by $$ A_{m}=\left[a_{1},a_{2},a_{3},...,a_{ T_{ N } } \right]^{ T }\in {  \mathbb{R} }^{ T_ {N} \times 1 }.$$

\noindent Our goal is to select segments from the video by Top-k highest importance prediction score as summarization set $\left\{ S \right\} _{k}$. So the problem can be defined as finding a ranking function $f\left( s \right) $ that can predict the segment importance score in video segments sequence $$ \hat { A_{m} } = f\left( V_{m} \right). $$
A loss function can be defined by mean-square error(MSE), $$L({ A }_{ m }\quad,\hat {A_{ m }} )=\frac { 1 }{ H\times T_{ N } } \sum _{ m=1,i=1 }^{ H,T_{ N } }{ \left( \left( A_{ m }^{ i }-f(V_{ m }^{ i }) \right) ^{ 2 } \right)  }, $$
\noindent where H is the number of videos in training set, our algorithm is to find optimal function $f$ that minimizes the overall loss: $$\hat { f } =\underset { R(f) }{ \quad argmin } \quad L(f), $$ in the prediction model space $R$.

Since $s_{i}$ lacks information from the whole video sequence as the input of $f$, we set ${V_{m}}'$  defined by $$ {V_{m}}' = V_{m} \oplus V_{s}, $$
$\oplus$ is the operator that concatenate each row of $V_{m}$ to $V_{s}$, $V_{s}$ is a $T_{N} \times vd $ matrix that represent $vd$ dimension whole video feature to each segment, which can also be taken as a learnable function of $V_{s}=\phi (V_{m})$. In this work, we set the same value for each row of $V_{s}$.

 We can also get sequence descriptors for one segment, like image frame, audio frame, a segment is defined by $$  S_{p}=\left[f_{1},f_{2},f_{3},\ldots,f_{T_{G}} \right]^{ T }\in {  \mathbb{R} }^{ T_ {G} \times fd }. $$ 

Another learnable function $V_{r}=\theta (S_{p})$ is used to map frames sequence features from $T_{G} \times fd$ to $1 \times sfd$ space. We combine image based feature $V_{r}$ and segment based feature $V{w}$ by learnable function $\varphi$
 to final segment-level feature $$V_{m}=\varphi (V{r},V{w}).$$
We can minimize $f$, $\phi$, $\theta$, $\varphi$ in one framework by minimizing
\begin{equation*}
\begin{split}
 L(f)&= L(f({ V_{ m } }')) \\
& = L(f(V_{ m }\oplus V_{ s })) \\ 
& = L(f(V_{ m }\oplus \phi (V_{ m }))) \\ 
& = L(f(\varphi (\theta (S_{p}), V_{w}  ) \oplus \phi (\varphi (\theta (S_{p}), V_{w}  )  ) )).
\end{split}
\end{equation*}

We design a video summarization network which consists of three subnetworks, SegNet for function $\theta$ and $\varphi$, VideoNet for $\phi$, and HighlightNet for $f$ as shown in figure 2.

As we utilize ImageNet features for each sampled video frame, SegNet combines all frame features' sequence to a frame-based vector. SegNet also fuses frame-based feature and segment-level feature together. The SegNet is designed to process either 2D or 3D frame sequence. For 2D features, we adopt temporal-convolution and pooling to squeeze temporal dimension. And for 3D features, we use three 3D convolution blocks. The 2D convolution blocks adopt “bottleneck” structure for spatial fully connected layers(figure 3). 

\begin{figure}[t]
\begin{center}
   \includegraphics[width=1.0\linewidth]{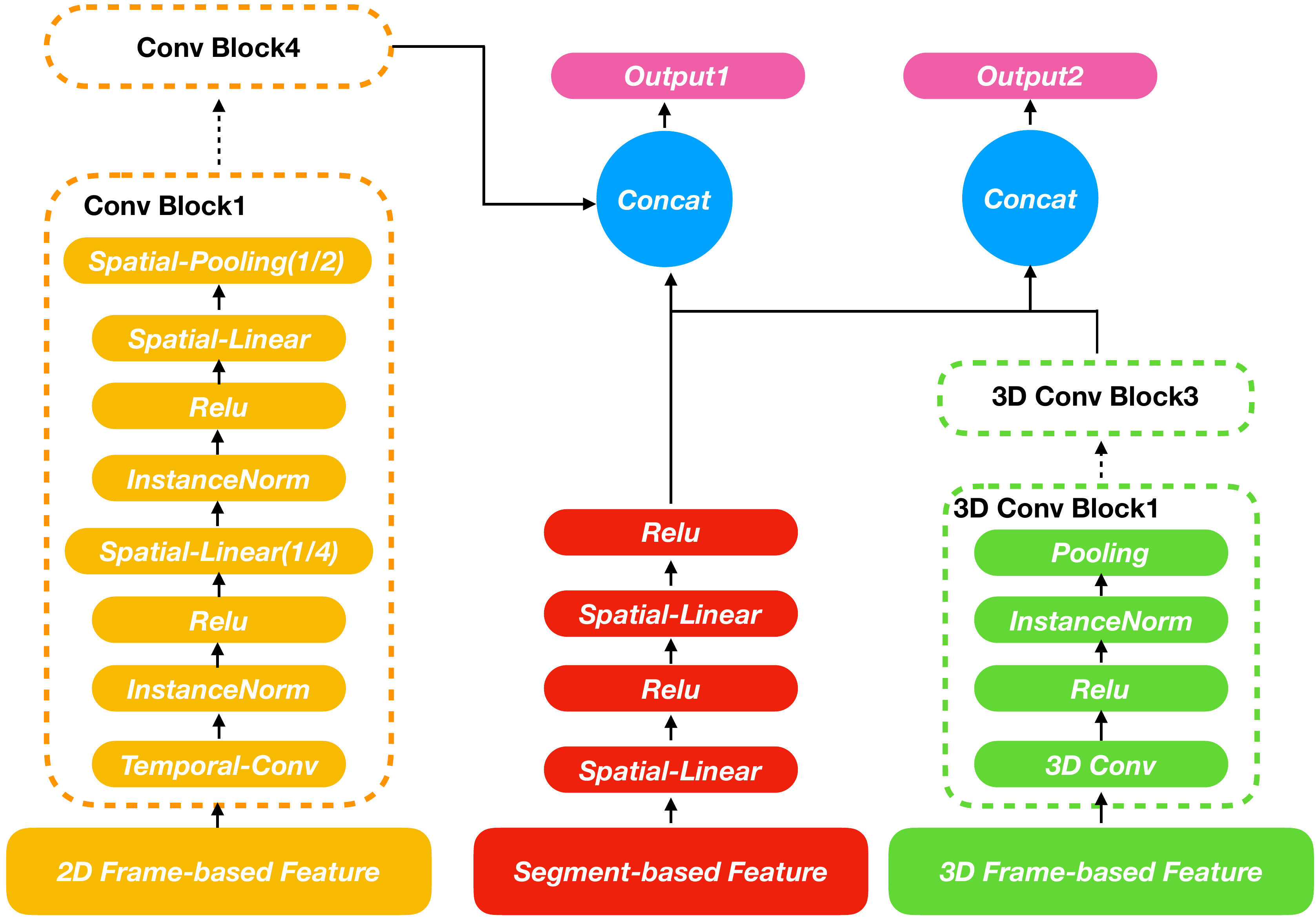}
\end{center}
   \caption{The SegNet structure details. The network is designed to process either 2D or 3D frame sequence. Only one output(output1 or output2) is taken as the segment embedding.}
\label{fig:long}
\label{fig:onecol}
\end{figure}

For each long video, the segments' feature sequence generated by SegNet are taken as inputs to VideoNet for video-level modelling. VideoNet use Bi-GRU for sequence encoding. The final GRU unit hidden layer, which is a fixed-length context vector, is taken as the output of VideoNet representing the video-level feature.

Each segment-level feature concatenates with the video-level feature of the same long video as the final segment representation. The representation contains not only the independent video segment information, but also the whole video sequence information, which is passed to highlight subnetwork to predict the segment importance score. Figure 4 shows the structure of highlight subnetwork.

\begin{figure}[t]
\begin{center}
   \includegraphics[width=0.8\linewidth]{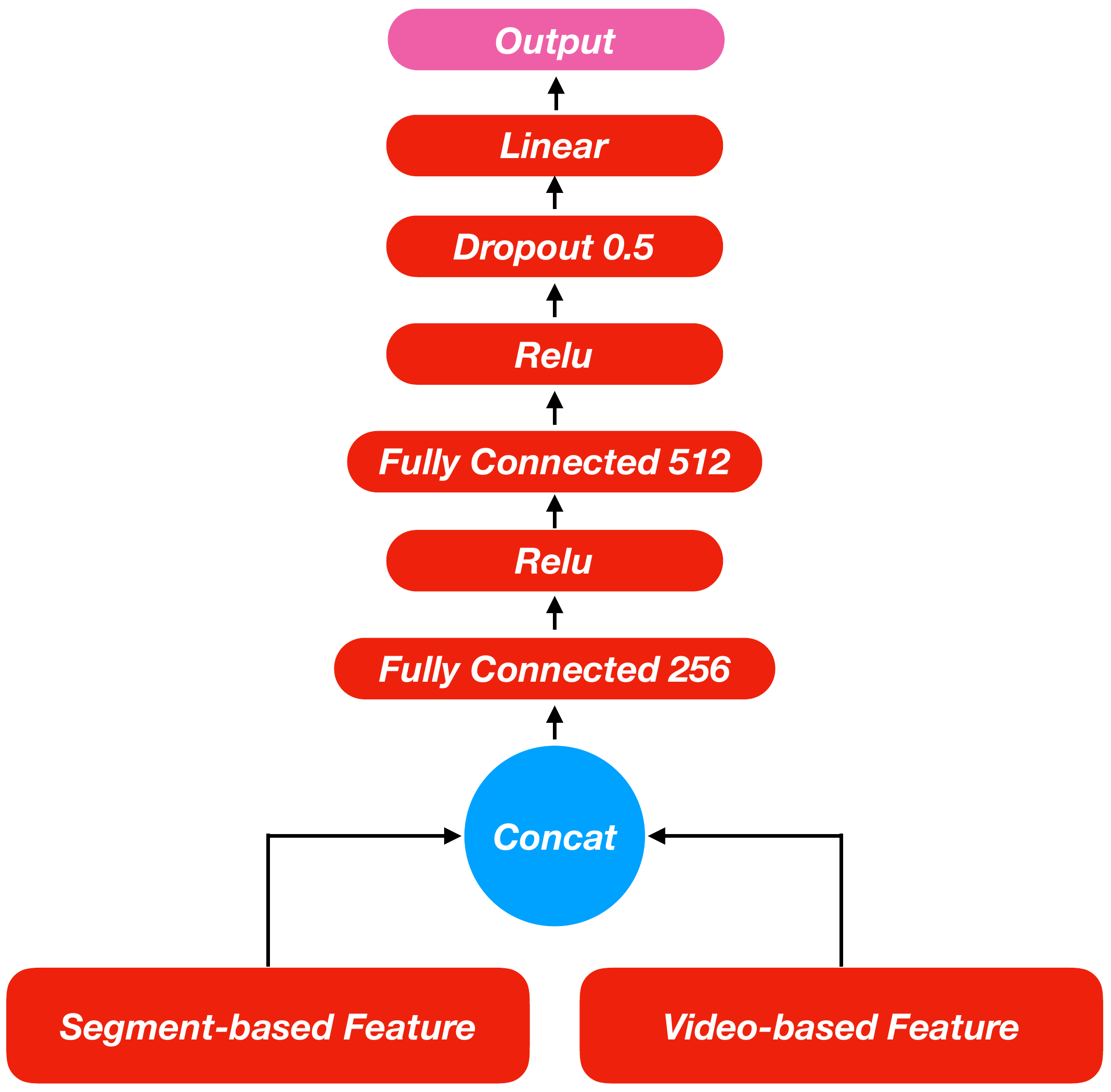}
\end{center}
   \caption{The HighlightNet structure details. The network is designed to predict the segment importance score by fusion features.}
\label{fig:long}
\label{fig:onecol}
\end{figure}

\subsection{Multi-task Learning and Data Augmentation for Video Summarization Network}

We consider an auxiliary self-supervised task to better modelling video sequence. Some segments are selected from one video by the fixed proportion before inputting to the network. We shuffle the selected segments and train the network to distinguish the odd-position segments as shown in figure 5. This operation assumes that a good video sequence encoder has the ability to model the right segments order. We control the difficulty of the task by only indicate the odd-position segments but not sort the shuffled sequence to right order. Parameters $\alpha$ and $\beta$ are used for adjusting the shuffled segments‘ proportion and the weight of self-supervised task in multi-task learning. The final loss can be defined by: $$ L(f)=L(f({ V_{ m } }')+\beta * \delta ( shuffle(\alpha, V_{ m }) )). $$

\noindent The advantages of applying multi-task learning are: 

   1. Learning several tasks simultaneously can suppress early-stage overfitting, by sharing the same representations.

   2. The auxiliary task is helpful in providing more useful information.

   3. Our method implicitly performs data augmentation since we shuffle the input video sequence, which may improve the robustness of the algorithm.

   4. This method utilize unlabelled data for video sequence modelling.

\begin{figure}[t]
\begin{center}
   \includegraphics[width=1.0\linewidth]{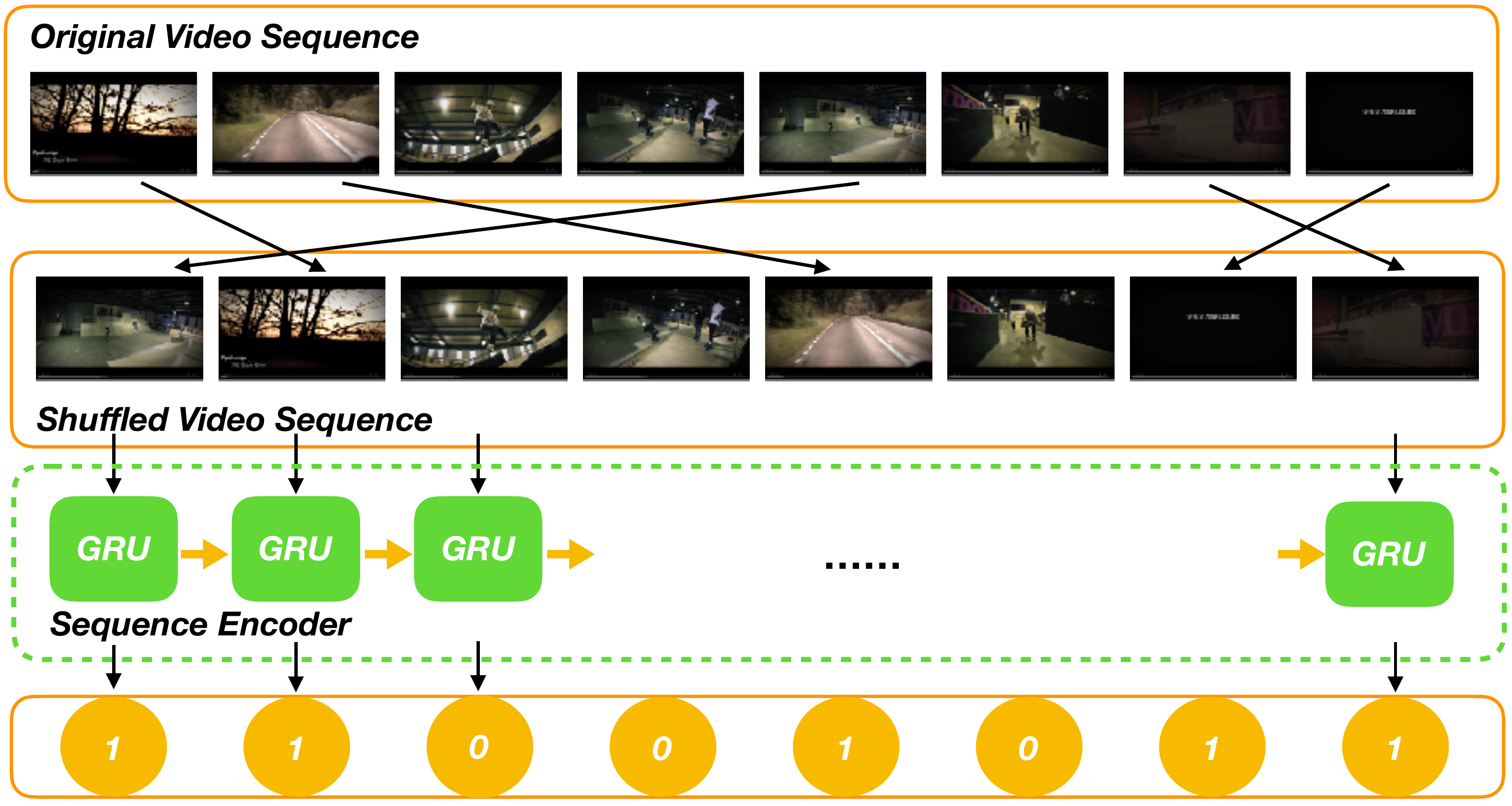}
\end{center}
   \caption{Self-taught learning for odd-position labelling. The input segment sequence are shuffled randomly by $\alpha$ ratio. The video encoder learn to find the odd-position segments in the sequence.}
\label{fig:long}
\label{fig:onecol}
\end{figure}

A data augmentation method is proposed which only choose a portion of information from each segment when modelling the whole video each time. After modelling all portion of segments, we average the embeddings from different portion to one vector(figure 6).

\begin{figure}[t]
\begin{center}
   \includegraphics[width=1.0\linewidth]{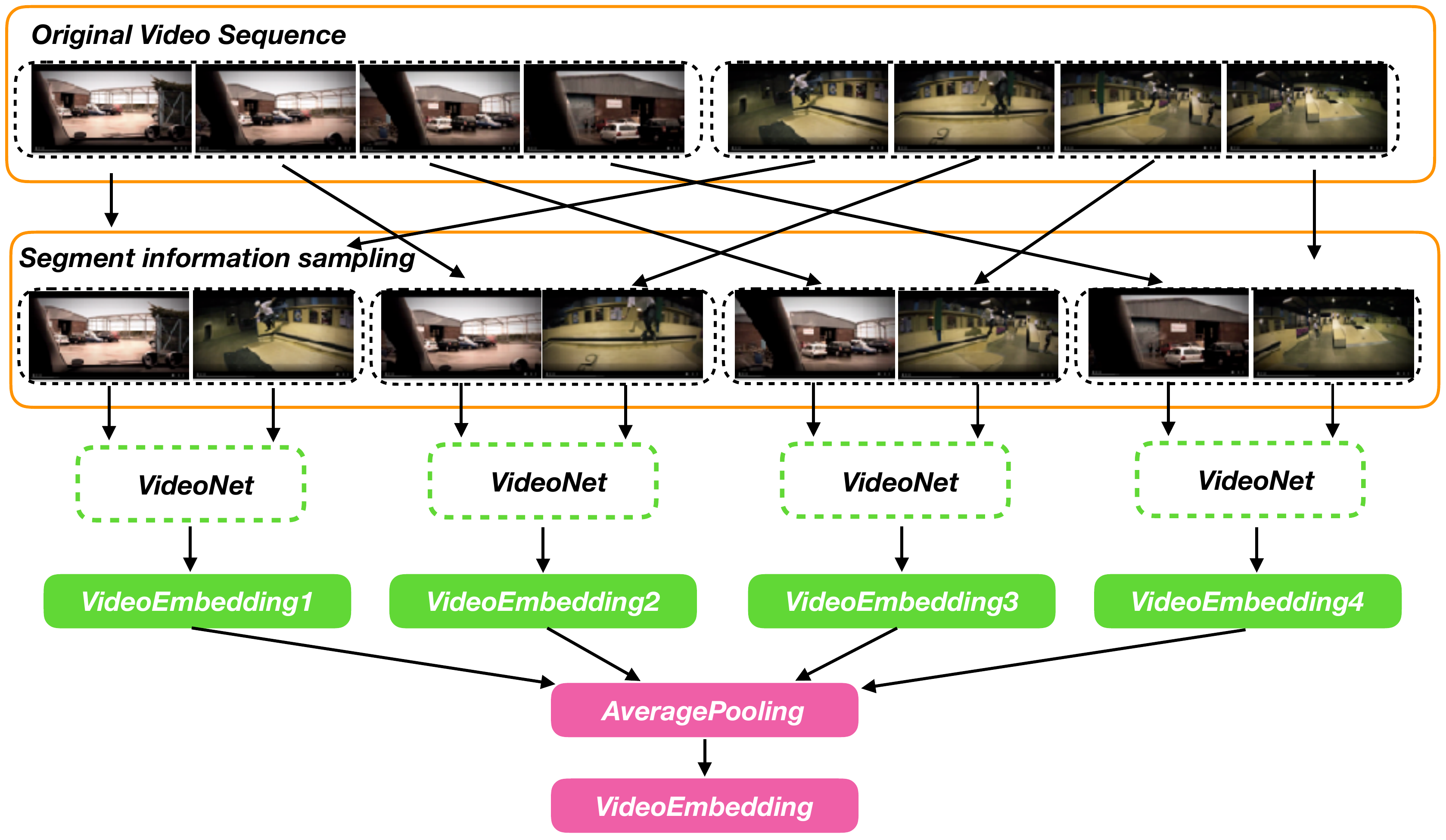}
\end{center}
   \caption{Data augmentation method. VideoNet only models a portion of each segment to video embedding. Multiple embeddings are combined by average pooling.}
\label{fig:long}
\label{fig:onecol}
\end{figure}

\section{Experiments}
In this section, we provide our comparison experiment on CoView 2019 dataset. The dataset consists of 1200 videos for training and 300 for testing. Videos are sampled from Youtube-8M, Dense-Captioning, and Video summarization dataset. Every video from the dataset is segmented into a set of 5-second-long segments and asked 20 users to annotate importance score and 99 action / 79 scene labels. The average of importance scores and the most voted action/scene labels are provided as ground-truth in segment-level. Before any experiments, we randomly split videos into training(1000) and validation (200) set based on original videos instead of five-seconds videos.

\subsection{Scene and Action Classification in Untrimmed Video}
For scene/action classification task, top-5 accuracy on validation results show in table 1. a. joint training I3D+non-local attention with two SoftMax branches(JT). b. independent I3D+non-local model(IT). c. independent I3D+non-local model with external action classification data(AED). d. ResNet-50 with external scene image classification data(SED). 

\begin{table}
\begin{center}
\begin{tabular}{|l|c|c|}
\hline
Method & Scene Top-5  & Action Top-5 \\
\hline
a (JT) & 82.76 & 81.72 \\
b (IT) & 86.92 & 87.77 \\
c (AED) & - & 84.86 \\
d (SED) & 82.06 & - \\
\hline
\end{tabular}
\end{center}
\caption{Top-5 accuracy of different models on validation.}
\end{table}

We train I3D+non-local with the pre-trained model on Kinetics, and the ResNet-50 pre-trained on ImageNet. The results indicate: 1. Although scene/action classification task may have inner connection, both joint training and ensemble failed to improve the results. 2. Pure 2D convolution-based network method is less accurate than 3D convolution-based network on both scene and action recognition task. 3. External data didn't help improve classification accuracy on validation set, but we still use it for potential model generalization enhancement.

\subsection{Video Summarization}
We investigate our summarization network on different parameter scales, visual and visual-audio feature inputs, single-task and multi-task learning, with and without data augmentation.

{\bf Evaluation Protocol.} The video summarization evaluation metric computes the sum of importance scores of selected segments comparing with ground truth summary segments. The importance score metric is defined as $$\frac { 1 }{ N } \sum _{ k=1 }^{ N }{ \frac { \sum _{ i=1 }^{ { N }_{ s } }{ ImportanceScore(k,\quad Sub_{ i }) }  }{ \sum _{ i=1 }^{ { N }_{ s } }{ ImportanceScore(k,\quad GT_{ i }) }  }  }, $$

\noindent where N is the number of test videos, $N_{s}$ is the number of summarized segments. $ImportanceScore(k, GT_{i})$ is the importance score from the i-th segment of the ground truth summary of the k-th video, and $ImportanceScore(k, Sub_{i})$ is the importance score of the i-th segment of the submitted summary for the k-th video. Importance Score is shared in both ground truth and submitted summary. $N_{s}$ is set to 6 for all videos, the top $N_{s}$importance segments are ground truth summary. We randomly choose $N_{s}$ segments for 10 times as the baseline of summary score, the mean value is 74.92\%.

{\bf Embedding Dimension.} We explore different embedding dimension for the network by regularizing the same length of segNet and videoNet outputs. For each segment, we sample 16 frames for image-based extractor and 32 frames for video-based extractor. We use GoogleNet trained on ImageNet to extract image-based feature and I3D+non-local blocks trained on Kinetics to extract video-based feature. Table 2 shows our results for different feature length from 512 to 64. The 256-feature length is slightly better.

\begin{table}
\begin{center}
\begin{tabular}{|c|c|c|}
\hline
Feature Length & SummaryScore \\
\hline
Dim = 512 & 81.47 \\
Dim = 256 & 81.81  \\
Dim = 128 & 81.49  \\
Dim = 64  & 81.19 \\
\hline
\end{tabular}
\end{center}
\caption{Summary score on different embedding dimension.}
\end{table}

{\bf Semantic Features Combination.} Image-based feature is extracted by image classifier. We select three types of image classifier: 1. ResNet-50 trained on CoView2019 scene classification task(R50\_s). 2. ResNet-152 trained on ImageNet(R152\_i). 3. GoogleNet trained on ImageNet(G\_i). Video-based feature is extracted by video classifier. We choose two video classifier: 1. I3D+non-local model trained on Kinetics(INK). 2. I3D+non-local model trained on CoView2019 action recognition task(INA). Table 3 shows different visual feature combination on feature length 256. Our model didn't benefit a lot from action and scene recognition task.

\begin{table}
\begin{center}
\begin{tabular}{|c|c|c|}
\hline
Feature Type & SummaryScore \\
\hline
G\_i+INK   & 81.81 \\
G\_i+INA   & 81.06 \\
R50\_s+INA & 80.83  \\
R152\_i+INA   & 80.99 \\
R50\_s+INK    & 81.45 \\
R152\_i+INK    & 81.45 \\
\hline
\end{tabular}
\end{center}
\caption{Summary score with different visual semantic feature combination.}
\end{table}

{\bf Multi-Task Learning.} Since the self-taught learning increases the learning difficulty, we adopt 3-layers bi-GRU on video net and feature length equals to 512 with R50\_s+INK features. Dropout and bottleneck structure are removed from segNet. The self-taught learning is firstly trained for two days using videos on CoView2019 with $\alpha=15\%$ shuffle ratio. We use the trained model, with 96.64\% accuracy on self-taught task and 78.6\% recall on odd order segments, as pre-trained model in multi-task learning setting. In joint training, we set $\alpha=0.02, \beta=1$, and $\alpha=0$ when evaluate on validation. Table 4 shows the comparison results, we evaluate three models: a. the baseline supervised learning without pre-trained model(sup\_no\_prt). b. supervised learning with self-taught pre-trained model(sup\_with\_prt). c. multi-task learning with self-taught pre-trained model(multi\_with\_prt). The results shows that multi-task setting is better than the baseline model with small margin.

\begin{table}
\begin{center}
\begin{tabular}{|c|c|}
\hline
Method & SummaryScore \\
\hline
a. (sup\_no\_prt)  & 80.69 \\
b. (sup\_with\_prt)  & 80.98 \\
c. (multi\_with\_prt)  & 81.64 \\

\hline
\end{tabular}
\end{center}
\caption{Summary score with Multi-Task learning.}
\end{table}

{\bf Data augmentation.} The basic network setting is the same as R50\_s+INK in "Semantic Features Combination" experiment, we get 0.29\% improvment from 81.45\% to 81.74\%.

{\bf Audio feature.} We extract audio features MFCC, Chroma-gram for each segment by using python audio processing package LibROSA. We concatenate the audio features to 5450-dimension segment-level features before inputing to SegNet. The audio fusion model builds bases on Multi-Task setting in "Multi-Task Learning" experiment. The result decrease from 81.64\% to 80.42\%, that may due to increased model complexity caused by larger feature dimension without important information.

{\bf Optical flow.} Optical flow as a low-level feature describing video motion, is shown as complementary information to RGB in action recognition task. Optical flow feature sequence are obtained by using Gunnar Farneback’s algorithm to extract 16 frames for each segment. We modified SegNet in Multi-Task setting to 4 layers 3D convolution for processing optical flow features on temporal and spectral. Same as audio feature, this modification decrease summary score from 81.64\% to 80.78\%.

{\bf Model ensemble.} We ensemble five models: base 256 feature length model(81.81\%), audio feature model(80.42\%), optical flow model(80.78\%), multi-task model(81.64\%), data augmentation model(81.74\%) by linear regression. Then we do model selection according to linear regression weights which direct us choosing base 256-length model and data augmentation model ensemble result(81.86\%) on validation as the final submission.

\section{Conclusion and Future Work}
In this paper, we propose a scalable deep neural network for video summarization in content-based recommender formulation, which predicts the segments importance score by considering both segment and video level information. Our work shows that data augmentation and Multi-Task learning is helpful in solving the limitation of dataset problem. To better understand the video content, we also perform action and scene recognition in untrimmed video with state-of-the-art video classification algorithm. We experiment with combinations of high-level visual semantic features, audio features and optical flow, and concluded that visual semantic features play the most important role in this summarization task.

There are some areas we can further explore to improve comprehensive video understanding in the future. First of all, we didn't benefit from action and scene connection in both recognition and summarization task, which leaves us room for better utilize this prior knowledge. Second, as we re-formalize video summarization in recommender framework, some state-of-the-art recommendation technologies can be introduced such as Collaborative Filtering, Factorization Machine, Wide \& Deep Learning and so on. Last but not the least, although CoView2019, to our knowledge, provides the largest public video summarization dataset, it is too small in the scale of deep learning. We need more, like users’ action when browsing video websites, video language description, and large-scale dataset to accomplish such complex task.

{\small
\bibliographystyle{ieee_fullname}
\bibliography{finalbib}
}
\end{document}